\documentclass{article}
\usepackage[a4paper, total={7in, 9in}]{geometry}

\usepackage{biblatex}
\usepackage[english]{babel}
\usepackage{authblk}

\usepackage{pifont}
\usepackage{xcolor}
\usepackage{algpseudocode}
\usepackage{algorithm}
\usepackage{graphicx}
\usepackage{booktabs}
\usepackage{amssymb}
\usepackage{hyperref}
\usepackage{caption}

\title{Kozax: Flexible and Scalable Genetic Programming in JAX}
\author[1*]{Sigur de Vries}
\author[1]{Sander W. Keemink}
\author[1]{Marcel A. J. van Gerven}
\date{}
\affil[1]{Department of Machine Learning and Neural Computing, Donders Institute for Brain, Cognition and Behaviour, Radboud University,  Nijmegen, the Netherlands}
\affil[*] {sigur.devries@ru.nl}

\begin{document}
\maketitle

\begin{abstract}
  Genetic programming is an optimization algorithm inspired by evolution which automatically evolves the structure of interpretable computer programs. The fitness evaluation in genetic programming suffers from high computational requirements, limiting the performance on difficult problems. Consequently, there is no efficient genetic programming framework that is usable for a wide range of tasks. To this end, we developed Kozax, a genetic programming framework that evolves symbolic expressions for arbitrary problems. We implemented Kozax using JAX, a framework for high-performance and scalable machine learning, which allows the fitness evaluation to scale efficiently to large populations or datasets on GPU. Furthermore, Kozax offers constant optimization, custom operator definition and simultaneous evolution of multiple trees. We demonstrate successful applications of Kozax to discover equations of natural laws, recover equations of hidden dynamic variables, evolve a control policy and optimize an objective function. Overall, Kozax provides a general, fast, and scalable library to optimize white-box solutions in the realm of scientific computing.
\end{abstract}

\section{Introduction}
Genetic programming (GP) is an evolutionary algorithm which automatically optimizes the structure of computer programs for input-output mapping~\cite{koza1994genetic}. GP does not require a fixed structure to be pre-selected for the solution, which allows the algorithm to flexibly discover general structures of solutions with less human bias. The computer programs are often represented by parse trees, consisting of functions and variables. GP discovers interpretable solutions that provide understanding about the data or model, and consequently GP has become one of the main pillars in the field of automated scientific discovery~\cite{wang2023scientific, bongard2007automated}. In scientific discovery, GP has been used in (re-)discovery of natural laws~\cite{schmidt2009distilling, la2021contemporary}, symbolic regression of dynamical systems~\cite{bongard2007automated, cao2000evolutionary}, learning symbolic control policies~\cite{hein2018interpretable,vries2024discovering,nadizar2024naturally} and evolving learning rules~\cite{jordan2021evolving, bengio1994use}.

Most such GP applications are based on separate implementations, each individually tailored to a specific task. Ideally, one unifying framework would exist that allows users to apply GP to their target problem. However, a major issue in GP is the high computational requirements needed to perform the fitness evaluation~\cite{harding2007fast}, especially when applied to difficult problems or large datasets. This complicates the development of a GP framework that both generalizes to arbitrary problems and runs efficiently. Variations of GP have been proposed that are computationally more efficient, such as Cartesian GP~\cite{miller2015cartesian} and linear GP~\cite{brameier2007basic}, however these variants have limitations in turn, such as reduced interpretability and inefficient evolution. Another approach for reducing computation time, is to improve the parallelization of the evaluation of different candidate solutions. Due to the inherently parallel nature of evolutionary algorithms, candidate solutions can be evaluated independently~\cite{harding2007fast}. Nonetheless, this remains difficult for GP, as the individual solutions in the population may have different structures and sizes.
\noindent
\begin{figure}
    \centering
    \includegraphics[width=0.4\linewidth]{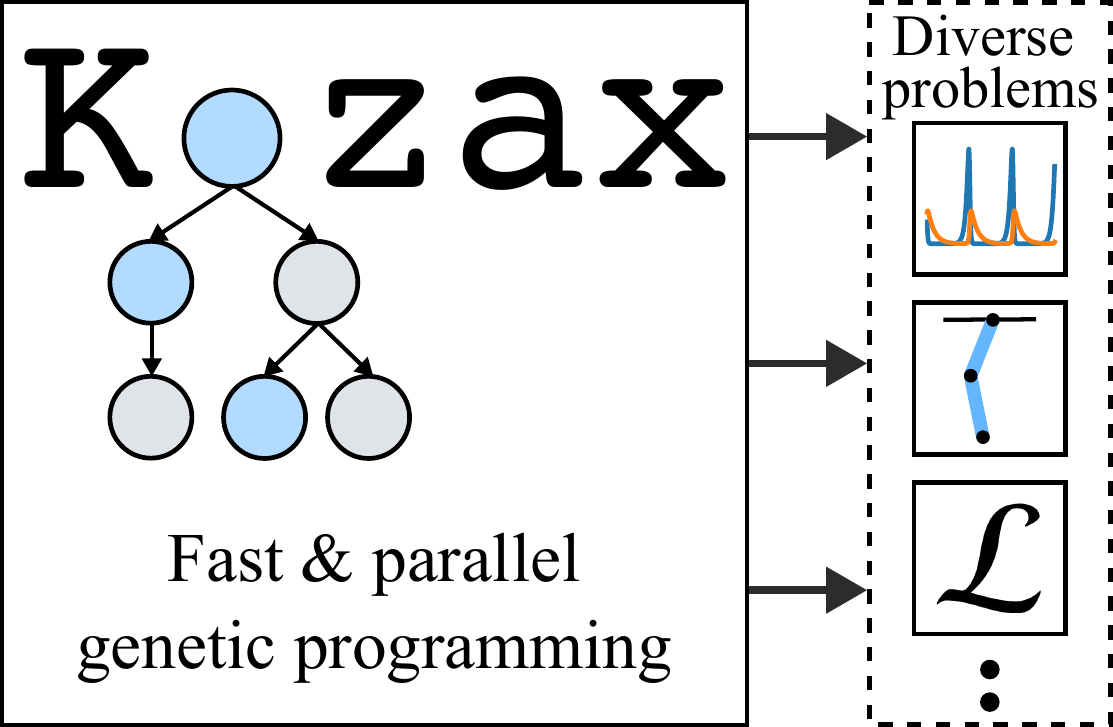}
    \vfill
    \captionof{figure}{We introduce Kozax as a general framework for genetic programming, utilizing JAX for fast and parallelizable computation. It allows for highly flexible problem and fitness function specifications, making Kozax applicable to diverse and complex tasks, including symbolic regression of dynamical systems, control policy optimization and objective function evolution for training neural networks.}
    \label{fig:logo}
\end{figure}

Previous attempts have been made to parallelize the fitness evaluation of GP on GPU. These ideas cover parallelizing the data points for each candidate~\cite{harding2007fast}, or distributing subpopulations over different GPUs~\cite{andre1998parallel, oussaidene1997parallel}, but in the end this still requires evaluation of subsets of candidate solutions sequentially. JAX~\cite{jax2018github} is an emerging Python framework for high-performance and efficient machine learning and numerical computing. JAX includes powerful features including vectorization of mathematical computation on CPU and GPU, just-in-time compilation and automatic differentation. Evosax~\cite{lange2023evosax}, EvoX~\cite{huang2024evox} and EvoJAX~\cite{tang2022evojax} provide implementations of genetic algorithms, evolution strategies, differential evolution in JAX with high parallelization. Although such algorithms are used to optimize numerical values instead of tree structures, it still shows the potential of JAX to reduce the computation time of GP.

In this paper we introduce Kozax\footnote{\url{https://github.com/sdevries0/Kozax}}, a JAX-based framework for GP to evolve symbolic expressions, named in honour of John R. Koza, the founder of GP. By representing the parse trees as matrices, the fitness evaluation is parallelized on CPU and GPU, showing reduced computation times when evaluating large populations or datasets. Like other libraries for GP, Kozax can be used for symbolic regression of laws and dynamical systems. However, Kozax can also be applied to other problems, and gives users much freedom to define the desired functionality for which symbolic expressions should be discovered, for example for control tasks or objective function optimization (Fig.~\ref{fig:logo}). Our results show that Kozax competes with other libraries on symbolic regression problems both in speed and performance, and extends to a larger range of complex problems. Overall, Kozax demonstrates to be a scalable, general implementation of the GP framework.

\section{Related Work}
Genetic programming (GP) was introduced by Koza in~\cite{koza1994genetic}, including many applications such as symbolic regression~\cite{koza1994genetic}, robot movement optimization~\cite{koza1992automatic}, planning, solving equations and finding a control strategy~\cite{koza1990genetic}. More recently, new tools have been developed that focus on symbolic regression using GP, either for dynamical systems or natural laws, including Eureqa~\cite{bongard2007automated, schmidt2009distilling}, DEAP~\cite{fortin2012deap}, Operon~\cite{burlacu2020operon} and PySR~\cite{cranmer2023interpretable}. Recent examples of applications of GP include control policy optimization~\cite{hein2018interpretable,vries2024discovering,nadizar2024naturally}, evolving learning rules~\cite{jordan2021evolving, bengio1994use} and objective functions~\cite{raymond2023fast}.

However, most GP frameworks were developed for specific applications, as there is still a lack of a unifying framework for GP. HeuristicLab~\cite{wagner2005heuristiclab} attempted to provide GP software that allows users to tune the fitness evaluation to their problems, but the software was limited by high computational requirements. DEAP~\cite{fortin2012deap} also offers flexibility in defining the fitness function and extending the optimization algorithm. However, DEAP does not use code compilation, therefore scales poorly when difficult fitness functions or large populations are evaluated. More specialized libraries for symbolic regression, such as Eureqa and PySR, apply tricks like partitioning and the finite difference method to the input data to improve the convergence to correct solutions. This requires the data to be provided in a specific format, focusing strictly on symbolic regression problems. Therefore, these libraries cannot be applied to other tasks involving numerical integration of differential equations or evaluating control policies in different environments without utilizing external methods. 

The specific data format required in Eureqa and PySR helps to reduce the runtime of fitness evaluation, which is typically the most time consuming process of GP. Parallelization of the fitness evaluation also improves the computation speed of GP. Previous work parallelized the evaluation of a single tree on many data points~\cite{harding2007fast}. However, this setup still requires the evaluation of different trees to be performed sequentially. A different approach is to evolve and evaluate different subpopulations on different processing nodes~\cite{andre1998parallel, oussaidene1997parallel}, which shows a linear relation between runtime and the number of subpopulations. To parallelize the evaluation of many solutions, general GP interpreters on GPU were developed~\cite{langdon2008simd, robilliard2009genetic, cano2014gpu, langdon2010many}, showing large speed improvements. More specific variants of GP, like geometric-semantic GP~\cite{trujillo2022gsgp, castelli2019gsgp} and stack-based GP~\cite{sathia2021accelerating, chitty2017faster}, have been made GPU-compatible, with speed improvements over the original version. 

In~\cite{nadizar2024naturally}, linear and Cartesian GP were implemented in JAX, allowing for parallelization of the fitness evaluation of all candidate solutions, as well as parallelization of the reproduction stage. Their implementation, however, was solely focused on learning control policies and does not allow users to apply the library to their own problems. NEAT is another direction in evolutionary computing that evolves the structure of small neural networks~\cite{stanley2002evolving}. Ref.~\cite{wang2025evogp} showed that with tensorization in PyTorch, the evaluation of the population can be sped up enormously on GPU compared to existing methods. Similarly, JAX can be used to parallelize fitness evaluation and significantly reduce the runtime of GP.

We developed Kozax with the intention to introduce a GP framework that provides both efficient and problem-independent fitness evaluation. Implemented in JAX, Kozax vectorizes the population, allowing for full parallelization of fitness evaluation, initialization and reproduction. Consequently, Kozax can run on either CPU and GPU, where the latter allows Kozax to scale the fitness evaluation to large datasets or populations. Furthermore, the functionality for fitness evaluation is not embedded in Kozax, therefore the fitness function can be adjusted by users for their problems of interest. This does mean that the computational efficiency of Kozax can be reduced when fitness functions are designed sub-optimally. To prevent users to define sub-optimal fitness functions, we provided multiple examples for various problems. The flexibility in fitness evaluation makes Kozax a general library for GP.

\begin{figure}[t]  
    \centering
    \includegraphics[width=0.9\linewidth]{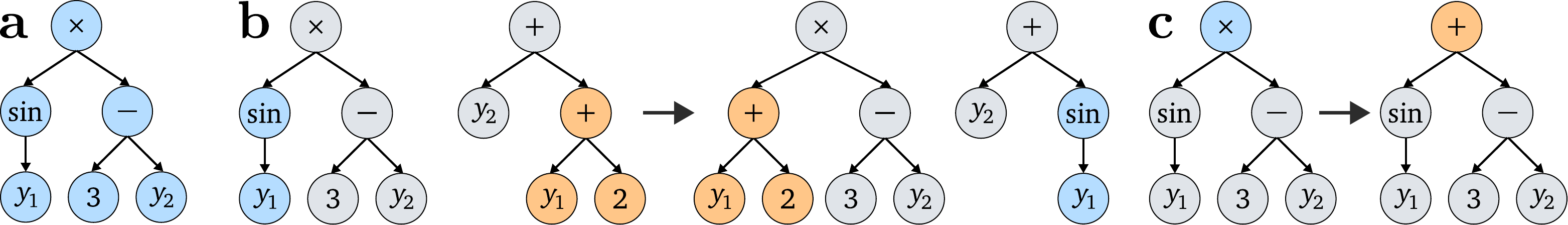} 
    \caption{\textbf{Overview of trees and reproduction in genetic programming.} (\textbf{a}) An example of the parse tree representation used in genetic programming. (\textbf{b}) An example of crossover on a pair of trees, where the blue and orange subtrees are swapped. (\textbf{c}) An example of mutation on a tree, where the operator in blue is replaced with the new operator in orange.}
    \label{fig: overview}
\end{figure}

\section{Genetic programming}\label{sec: GP}
GP is a variant of evolutionary algorithms that evolves the structure of computer programs~\cite{koza1994genetic}. In Kozax, we focus on evolving symbolic expressions represented by parse trees, a subset of computer programs. Parse trees consist of mathematical operators as interior nodes and variables and constants as the leaf nodes. Parse trees are executed recursively, where child nodes have to be computed before their parent node. An example of the parse tree representation is presented in Figure~\ref{fig: overview}a.

In GP, a population of solutions is optimized on a specified task through stochastic optimization. See Algorithm~\ref{alg: GP} for an overview of the GP algorithm. To evaluate a candidate solution, it is tested on a problem to compute a fitness score. The performance of a candidate solution is expressed by a fitness score, computed with a fitness function. The fitness scores are used to select individuals for reproduction, where fitter individuals produce more offspring.

In every generation a new population is evolved, consisting of new solutions that are generated by reproducing existing solutions with crossover and mutation. In crossover, the genotype of two individuals is combined to produce new solutions. In both parents a random node is selected, after which these nodes and the corresponding subtrees are swapped. An example of the crossover operator is shown in Figure~\ref{fig: overview}b. Mutation is applied to a single individual to generate one offspring. Many aspects of trees can be mutated, for example changing, deleting or adding operators, replacing subtrees or changing variables or constants. Figure~\ref{fig: overview}c shows an example of mutation, in which an operator is changed to a different type of operator.

\begin{figure}[H]
\begin{minipage}[t]{0.48\linewidth}
\begin{algorithm}[H]
\textbf{Input} Number of generations $G$, population size $N$, elite percentage $E$, fitness function $F$
\begin{algorithmic}[1]
    \State {\em initialize} population $P$ with size $N$
    \For{$g$ in $G$}
        \State {\em evaluate} each individual in $P$ on $F$
         \State offspring $O$ $\longleftarrow \varnothing$ 
         \State append fittest $E$ of $P$ to $O$
         \While{size($O$) $<$ $N$}
            \State select parents $p$ from $P$
            \State children $c$ = {\em reproduce}($p$)
            \State append $c$ to $O$
         \EndWhile
        \State $P$ $\longleftarrow$ $O$
    \EndFor
    \State \Return fittest individual in $P$
\end{algorithmic}\caption{\textbf{Genetic programming algorithm}}\label{alg: GP}
\end{algorithm}
\end{minipage}
\hfill
\begin{minipage}[t]{.5\textwidth}
\vspace{0.42cm}  
\begin{small}
    \centering
\begin{tabular}{lccc}
\toprule
\textbf{Feature} & \textbf{PySR} & \textbf{DEAP} & \textbf{Kozax} \\
\midrule
Custom operators & \checkmark & \checkmark & \checkmark \\
Custom fitness function & \checkmark & \checkmark & \checkmark \\
Pareto front & \checkmark & \checkmark & \checkmark \\
Symbolic constraints & \checkmark & -- & --\\
Simplification & \checkmark & -- & --\\
Multi-objective optimization & -- & \checkmark & -- \\
JIT compilation & \checkmark & -- & \checkmark \\
Constant optimization & \checkmark & -- & \checkmark \\
Flexible tree definition & -- & \checkmark & \checkmark \\
Different tree classes & -- & -- & \checkmark \\
Runs on GPU & -- & -- & \checkmark \\
\bottomrule
\end{tabular}
\captionof{table}{Comparison of algorithmic and software features between PySR, DEAP and Kozax. A check mark indicates that this feature is included in the library, while a dash shows that this feature is currently missing.}
    \label{tab:features}
    \end{small}
\end{minipage}
\end{figure}

\section{Kozax}
Kozax is a GP library to efficiently optimizes the structure of symbolic expressions, implemented in JAX~\cite{jax2018github}. In this section, the major features of Kozax are described in more detail. Afterwards, we will explain the vectorized representation of solutions adopted in Kozax, and how this representation allows for parallelization of fitness evaluation.


\subsection{Features}
In Table~\ref{tab:features} we compare the most important features of Kozax to PySR~\cite{cranmer2023interpretable}, a highly specialized library for symbolic regression, and DEAP~\cite{fortin2012deap}, a customizable GP library. PySR and DEAP have some features that are currently missing in Kozax, such as placing constraints on symbolic operators, simplification of solutions during evolution and multi-objective optimization. In the remainder of this section, we will describe the major features integrated in Kozax.

First of all, Kozax makes use of just-in-time (JIT) compilation to minimize computational overhead and improve the computation speed. The fitness evaluation and reproduction functionality are compiled, resulting in large speed improvements, as the complete population is evaluated and evolved in parallel. Furthermore, Kozax allows users to define custom operators and fitness functions, which remains compatible with JIT compilation. During evolution, Kozax keeps track of the best solutions in a Pareto front. The Pareto front stores the best solution in terms of fitness at every complexity level, and finally includes solutions only when the fitness is improved over less complex solutions. Currently, expressions are only simplified when they are displayed, but simplification does not occur during evolution.

\begin{figure*}[!t]
    \centering
    \includegraphics[width=\linewidth]{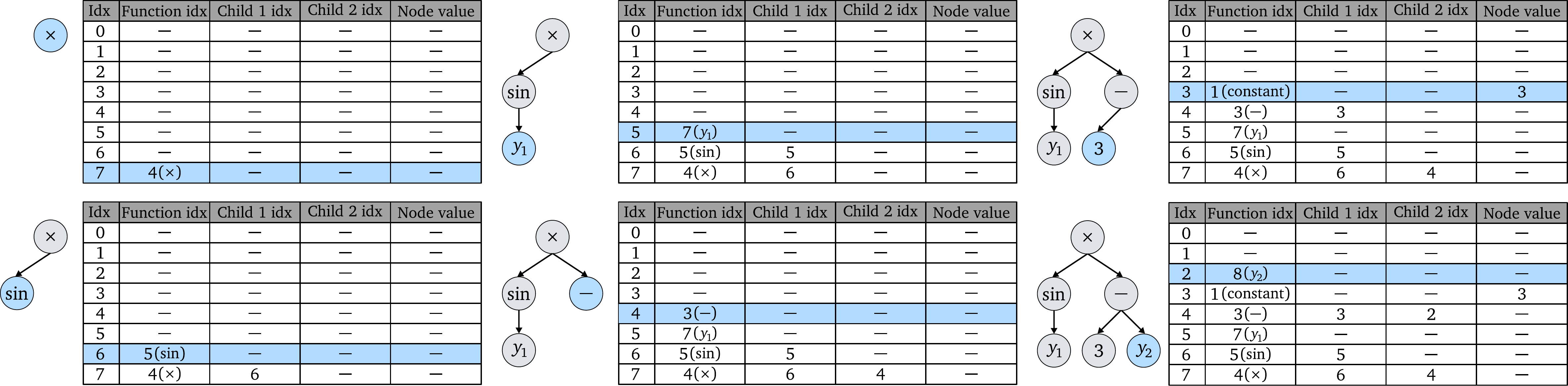}
    \caption{\textbf{Step-by-step mapping of a tree to a matrix.} The node added in the tree and the corresponding row in the matrix are marked blue at every step. The references to child nodes are added to the relevant rows once the child node itself has been added to the tree. The matrix is inverted, as the execution starts with the leaf nodes.}
    \label{app: matrix}
\end{figure*}

An important improvement to the standard GP algorithm is to integrate an external method for optimization of the constants in individual solutions~\cite{topchy2001faster}. Kozax provides gradient-based optimization, based on automatic differentiation in JAX, and a simple version of a genetic algorithm to optimize constants. Kozax offers these two methods for constant optimization, as either method may be more effective for balancing computation time and performance in specific tasks. To reduce the computational load, users can specify the number of solutions to apply constant optimization to. When using gradient-based optimization, the number of epochs for each optimized solution can be defined, and when using genetic algorithms, both the number of iterations and the population size are tunable. These factors together result in the number of constant optimization steps.

One major advantage that differentiates Kozax from other GP implementations, is the ability for users to completely define the functionality for which trees are evolved. This includes when a tree is evaluated in the fitness evaluation, what its inputs are and where the output of a tree is used. A related feature is the ability to optimize multiple trees in Kozax, with an option to provide different operator and variable sets for individual trees. These two options allow Kozax to be applied to problems other than standard symbolic regression tasks, such as control policy optimization or inferring equations of hidden variables without requiring external methods.

Lastly, Kozax can parallelize the fitness evaluation and candidate evolution on both CPU cores and on GPU, even allowing for distribution over multiple GPUs. On GPU, Kozax scales the fitness evaluation to large populations or datasets efficiently, where high computation time would be obtained on CPU. Altogether, these features make Kozax a general, high-performing and scalable GP library.

\subsection{Tree representation}
Kozax implements the GP algorithm using JAX~\cite{jax2018github}. Typical applications of GP include symbolic regression, policy optimization and other problems that benefit from interpretable solutions. As these are challenging problems, being able to evaluate large populations efficiently is advantageous. JAX offers features for faster computation, high scalability and high-performance machine learning, like vectorization of functions, just-in-time compilation and automatic differentiation. Accordingly, Kozax makes use of these features to improve runtime, scale to large populations and evolve accurate solutions.

Parallelization of different tree structures in GP is not trivial, as the trees in a population have varying structures. To this end, the trees are represented as matrices with a fixed size in Kozax. The size of the matrix can be selected by users, balancing computational efficiency and tree complexity. In Cartesian and linear GP the matrix representation is inherent and has already been implemented in JAX~\cite{nadizar2024naturally}. However, JAX currently lacks a general implementation of tree-based GP. Figure~\ref{app: matrix} presents a step-by-step mapping of a tree to a corresponding matrix. Figure~\ref{app: matrix} is purely illustrative, as mapping to matrix form produces computational overhead and therefore solutions are always represented as matrices in Kozax. Each row in the matrix represents a node in the tree, limiting the number of nodes in a tree to the matrix size. A row defines a node with an integer representing the function of the node, the node indices of the children of this node and the computed value of the node, which is set to 0 before execution. Before initialization, Kozax maps the set of operators and variables to unique indices to cover the possible values of nodes. Constants are represented with a function index set to 1, and the value of the constant is directly stored in the solution column. Empty rows are represented with zero and nodes without children do not have references to other nodes.

\begin{figure*}[!t]
    \centering
    \includegraphics[width=\linewidth]{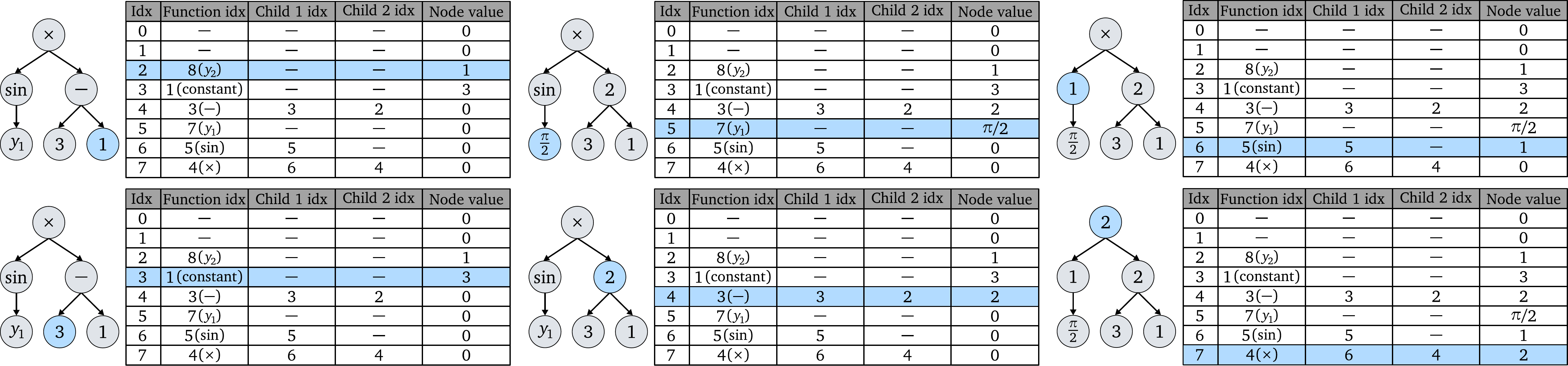}
    \caption{\textbf{Processing of a matrix in Kozax.} The matrix is iteratively solved with the input values: $y_1 = \frac{\pi}{2}$ and $y_2 = 1$. A blue row in the matrix corresponds to a blue node in the tree, where the computed value of the node is stored in the last column. The final value of the tree is obtained by taking the stored value in the last row.}
    \label{fig: solution}
\end{figure*}

To execute a tree and compute the output given inputs, the rows in the matrix are solved iteratively.  During evaluation, the computed value of a node is stored in the final column, which is available to nodes higher in the tree. To compute the value of a certain node, the corresponding function is executed with the value of the child nodes as inputs, as well as the values of the input variables. To parallelize the execution of trees, the full function and variable library is evaluated given inputs, and the relevant value is selected given the function index of the node. If the node represents a constant, the value of the constant is stored. If the node represents a variable, it will store the relevant value from the provided input variables. If the function represents an operator, it will compute the value given the inputs and store this. An empty node will return 0, but is never referred to by other nodes. The matrix is defined with the root node as the last row, as solving the tree requires that the values of child nodes are computed before the parent node is executed. The final outcome of the tree is obtained by retrieving the stored value of the last row. An example of solving the computation in a matrix in Kozax is displayed in Figure~\ref{fig: solution}.

\begin{figure*}[!t]
    \centering
    \includegraphics[width=0.6\linewidth]{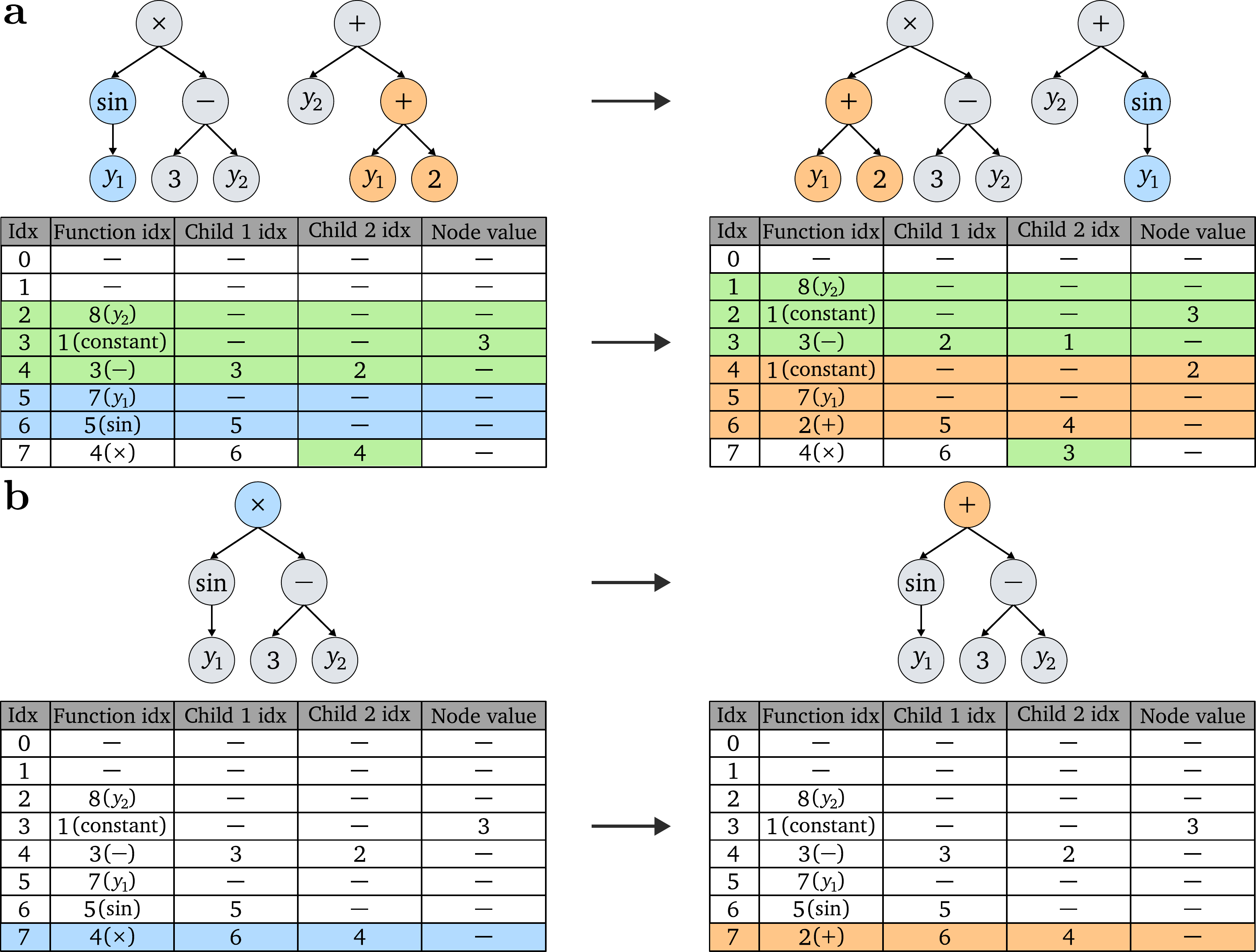}
    \caption{\textbf{Evolution of new trees in Kozax.} (a) Crossover applied to a pair of trees, producing two new trees. A random node is selected in both trees and the corresponding subtrees are swapped, indicated by the blue and orange subtrees. The matrix shows the representation of the left tree before and after crossover, where the blue and orange rows correspond to the removed and added subtrees respectively. The green cells show the nodes that remain in the tree, but of which the position or child indices have been changed accordingly. (b) Mutation is applied to a tree to evolve a new tree. In this example, the root node changes from a multiplication to an addition. The matrix representation is shown before and after mutation in blue and orange respectively.}
    \label{fig: evolution}
\end{figure*}

With this representation, all trees can be evaluated simultaneously. Kozax also performs initialization and reproduction in matrix space, therefore these stages are also vectorized to further reduce computation speed. Unlike grammar evolution~\cite{o2001grammatical} or Cartesian GP~\cite{miller2015cartesian} that make use of matrices or grid-like structures, Kozax adheres to the tree hierarchy in standard GP throughout evolution. This means that the matrices have to be adjusted accordingly when trees are evolved. Figure~\ref{fig: evolution}a shows how crossover is applied to a pair of trees in Kozax. Old rows resembling a subtree in the current tree are replaced with rows representing a subtree from the other parent. Additionally, rows may have to be moved and index references have to be updated because the size of the subtree may change. In mutation, a single row is adapted to match the mutated node, as shown in Figure~\ref{fig: evolution}b. For other mutation types, similar functionality is implemented to evolve matrices.

\section{Results}
\begin{table}[!t]
\begin{small}
    
    \begin{tabular}{lllll}
    \toprule
    \textbf{Experiment} & \textbf{Generations} & \textbf{Population size} & \textbf{CO steps} & \textbf{Operators}\\
    \midrule
       Kepler's third law & 100 & 1000 & 25000 & $+$, $-$, $\times$, $\div$, power\\
       Newton's law of universal gravitation &100& 1000 & 25000 & $+$, $-$, $\times$, $\div$, power\\
       Bode's law & 100 & 1000 & 25000 & $+$, $-$, $\times$, $\div$, power\\
       Fully observable LV equations & 100 & 1000 & 25000 & $+$, $-$, $\times$, $\div$, power\\
        Partially observable LV equations & 100 & 2000 & 100000 & $+$, $-$, $\times$ \\
       Acrobot & 50 & 500 & 0 & $+$, $-$, $\times$, $\div$, power, $\sin$, $\cos$\\
       Objective function & 50 & 250 & 0 & $+$, $-$, $\times$, $\div$, power, $\log$, $\exp$\\
       \bottomrule
    \end{tabular}
    \centering
 \caption{Hyperparameters used in each experiment. CO steps refers to the number of constant optimization steps performed at each generation in Kozax. The LV equations represent the Lotka-Volterra dynamics.}
    \label{tab:hyperparam}
    \end{small}
\end{table}

\subsection{Experiments}
We tested Kozax in seven experiments to demonstrate its ability to find accurate solutions for a variety of problems. We chose to compare Kozax with PySR~\cite{cranmer2023interpretable} and DEAP~\cite{fortin2012deap}, as they are popular Python libraries for genetic programming. 
The experiments and hyperparameters used in each experiment are presented in Table~\ref{tab:hyperparam}. The set of operators was empirically determined for each experiment, which can easily be adjusted as Kozax accepts custom operators. The number of generations and population size are used in PySR, DEAP and Kozax, but the number of constant optimization steps is only relevant in Kozax. The experiments include symbolic regression of laws and dynamic systems, optimizing a symbolic control policy and evolving a loss function. In the symbolic regression experiments, the genetic algorithm is used for constant optimization in Kozax. The three libraries are evaluated on ten different seeds, which influence both the data generation and the initial population. Results are shown in Table~\ref{tab:comparison}. The results are reported as the average fitness of the best solutions found in 10 seeds with the standard deviation around the mean, and the average complexity of the best solutions. The fitness functions in each experiment are explained in the relevant sections. The raw data and code to reproduce the results are available at \url{https://github.com/sdevries0/kozax_paper}.

\subsubsection{Law discovery}
The first three experiments entail symbolic regression of Kepler's third law~\cite{kepler1997harmony}, Newton's law of universal gravitation~\cite{newton1987philosophiae} and Bode's law~\cite{bonnet1781contemplation}. The data of Kepler's third law and Bode's law are based on true observations. The data for Newton's law of universal gravitation is generated according randomly sampled values of the input variables, and noise is added to the output variable. A symbolic expression receives the inputs corresponding to each law, which are mapped to a predicted value. The fitness function is computed as the absolute mean error between the predictions and the targets.

PySR is able to evolve the correct equations for each of the three laws, as the fitness is always among the best of the three methods. DEAP obtains the highest fitness on all three laws, and the resulting expressions have high complexity compared to the expressions evolved with the other two methods. A possible explanation the poor performance of DEAP is the lack of constant optimization in the trees. Kozax obtains comparable performance to PySR on two laws, but the average fitness is higher than PySR for Kepler's third law. Overall, Kozax shows to compete with PySR, a highly specialized tool for symbolic regression.

\begin{table}[!t]
\centering
  \begin{small}
    \begin{tabular}{lcccccccc}
    \toprule
     \textbf{Experiment}& \multicolumn{2}{c}{\textbf{PySR}} & &\multicolumn{2}{c}{\textbf{DEAP}} &&\multicolumn{2}{c}{\textbf{Kozax}}\\
    \cline{2-3}\cline{5-6}\cline{8-9}& \textbf{Fitness} & \textbf{Size} &  & \textbf{Fitness} & \textbf{Size} & & \textbf{Fitness} & \textbf{Size} \\
    \midrule
       Kepler's third law & $\textbf{2.00}\pm0.10$&$5.2\:\:\pm0.6$ &&$9.69\:\:\:\pm9.19$&$83.1\pm26.0$& &$5.15\:\:\pm5.98$&$7.8\:\:\pm2.2$\\
       Newton's law & $\textbf{0.15}\pm0.21$&$10.2\pm3.3$&&$1.12\:\:\:\pm0.21$& $50.0\pm19.1$&&$\textbf{0.14}\:\:\pm0.19$&$9.6\:\:\pm1.3$\\
       Bode's law & $\textbf{0.07}\pm0.00$&$7.0\:\:\pm0.0$&& $0.28\:\:\:\pm0.11$& $27.6\pm  9.7\:\:$&&$ \textbf{0.07}\:\:\pm0.00$&$7.0\:\:\pm0.0$\\
       Fully observable LV &$\textbf{0.00}\pm0.00$ &$14.0\pm0.0$&&$0.47\:\:\:\pm0.20$&$26.1\pm4.4\:\:$&&$0.01\:\:\:\pm0.01$&$16.2\pm1.4$\\
        Partially observable LV & -- & \:\:--&&\:\:\:--&--&&$\textbf{0.24}\:\:\pm0.30$&$16.8\pm2.0$ \\
       Acrobot & -- &\:\:--&&$\textbf{0.32}\:\:\:\pm0.00$&$15.0\pm2.3\:\:$&&$\textbf{0.33}\:\:\pm0.01$&$9.3\:\:\pm1.8$\\
       Loss function (small) & -- & \:\:-- && $-0.72\pm0.02$&$3.4\:\:\pm2.7$\:\:&&$\textbf{-0.85}\pm0.15$&$8.9\:\:\pm4.3$\\
       Loss function (big) & -- & \:\:-- && \:\:\:-- & -- & &$\textbf{-0.98}\pm0.00$&$10.3\pm1.6$\\
       \bottomrule
    \end{tabular}
    \caption{Results of PySR, DEAP and Kozax on a set of experiments. Each experiment was repeated for 10 seeds, and the mean and standard deviation of the fitness is shown. Furthermore, the average complexity of the best solution, defined in the number of nodes, is shown with the standard deviation. The best methods in each experiment are marked bold. A dash means that an experiment was not possible to conduct given the library without involving external methods.}
    \label{tab:comparison}
    \end{small}
\end{table}

\subsubsection{Symbolic regression of dynamical systems}
The next experiment is symbolic regression of the Lotka-Volterra equations~\cite{goel1971volterra}, a dynamical system governing the population of preys and predators. As methods such as PySR and SINDy make use of the finite difference method, we opted to use this method in DEAP and Kozax as well. One trajectory of the Lotka-Volterra dynamics is integrated, given a randomly sampled initial condition. Afterwards, the true derivative of the state is computed for every time step. Symbolic regression is then performed with the states as the inputs and the derivatives as the target outputs. The fitness function is again the mean absolute error between the predicted and true derivatives of the state. PySR and Kozax are able to successfully rediscover the equations of both the prey and the predator, as the fitness approximates zero for both libraries. Again DEAP performs worse, both in terms of fitness and solution complexity.

The finite difference method allows symbolic regression of dynamical systems with PySR. However, this only works when all dimensions of the system are observed. In the fifth experiment, only the prey is observed in the Lotka-Volterra model, which requires GP to find two equations and integrate them as a system of ordinary differential equations. The evolved equations are integrated from the true initial condition, and the fitness is computed as the mean absolute error between the predicted and true prey population, plus an additional penalty when either population is negative at any timepoint. This is not possible in PySR without involving external methods, but in DEAP and Kozax the fitness function can be adjusted to integrate differential equations of unobserved variables. However, the integration of the differential equations is too expensive to perform for each candidate in DEAP, and consequently this experiment could not be performed using DEAP. In Kozax, the parallelization of the integration of differential equations on GPU reduces the required computation massively compared to DEAP. Kozax obtains poorer fitness than when data of both prey and predator is observed, but still the fitness and complexity of solutions show that it can substantially recover the equations.

\subsubsection{Symbolic control policy optimization}
In this experiment, a symbolic control policy was evolved to solve the acrobot swing-up task~\cite{sutton1995generalization}. The fitness function is a sparse reward function, returning the first time point at which the swing-up was satisfied, divided by the total length of the trajectory, which is fixed at 200 seconds. The proposed policies return continuous values, but these values are mapped to [-1, 1] given the sign of the policy's output. A policy observes the sine and cosine of the angles of the first and second link, and the angular velocities of the two links. The acrobot task is simulated for a fixed number of time steps using Gymnasium~\cite{towers2024gymnasium}. The efficiency of Kozax benefits of an implementation of the acrobot in JAX, which is provided in Gymnax~\cite{gymnax2022github}.

PySR could discover a symbolic policy by distilling a pre-trained black-box policy, however the accuracy heavily depends on the quality of the teacher policy, and information might be lost in the distillation stage. Therefore, it is advantageous to optimize a symbolic policy through direct interaction with the control environment. This means that PySR cannot be directly used to learn control policies. In DEAP and Kozax, it is possible to evaluate solutions on control tasks directly. Both DEAP and Kozax are able to consistently solve the acrobot, while Kozax finds solutions with lower complexity.

\subsubsection{Evolving an objective function}\label{sec: loss}
Learning the objective function is a direction within the field of meta-learning that can improve convergence speed and robustness~\cite{bechtle2021meta}. Previously, GP has been used to learn a symbolic expression for the loss function in various problems~\cite{raymond2023fast}. In the final experiment, a neural network is trained with backpropagation on the binary XOR classification problem using the evolved loss function. The input data consists of uniform samples between zero and one in two dimensions, and the target is zero when both dimensions of the input are either smaller than 0.5 or larger than 0.5, and one otherwise.

The loss function is tested on different batches of data to prevent overfitting of the loss function. For every candidate, a neural network is randomly initialized with two hidden layers, each with 16 hidden neurons. The activation functions of the two hidden layers are $\tanh$, and the final layer applies a sigmoid to produce logits. The neural networks are trained for 100 epochs with 50 data points per epoch, using an Adam optimizer with a learning rate of 0.01. The evolved loss function is applied to individual pairs of neural network predictions and targets, after which the loss is computed by averaging over the complete batch.  After training, the final neural network is applied to unseen data, and the test accuracy is computed given the predicted labels. The negative test accuracy is used as the fitness of the candidate loss function. The results are indicated in Table~\ref{tab:comparison} as loss function (small). As parallelizing the training of neural network benefits from running on GPU, Kozax can also be tested given a large dataset. Here, four neural networks are trained for every candidate function, with 500 epochs per network and 100 data points per epoch, and the population size is increased to 1000. The results of the larger dataset are presented by loss function (big) in Table~\ref{tab:comparison}.

During objective function optimization, the target outputs for fitness evaluation are not available. Therefore, PySR is incompatible with the task in this experiment. However, the fitness evaluation in DEAP and Kozax can be adjusted to learn the objective function to train neural networks. With the smaller dataset, Kozax already outperforms DEAP in terms of fitness of the best evolved solutions. The large dataset was too computationally intensive to run with DEAP, but Kozax could be applied and reaches an average test accuracy of 98.3\%.

\subsection{Runtime analysis}
\begin{figure}[!b]
    \centering
    \includegraphics[width=0.9\linewidth]{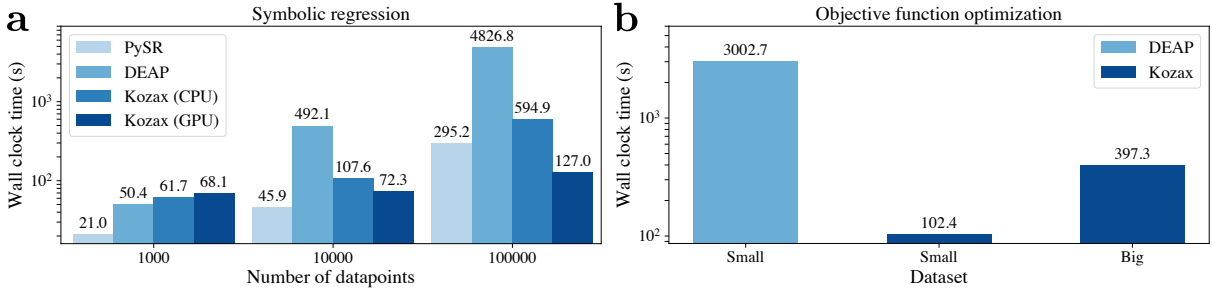}
    \caption{\textbf{Runtime analysis of PySR, DEAP and Kozax.} (a) The wall clock runtime is measured of symbolic regression with PySR, DEAP and Kozax to complete a fixed number of generations, given different dataset sizes. The runtime of Kozax is measured when running either on CPU and GPU. The wall clock time shows the average over three runs. (b) The wall clock runtime is measured of DEAP and Kozax in the objective function optimization experiment. Both DEAP and Kozax are tested on a small dataset, but Kozax is also tested on a large dataset}
    \label{fig:time}
\end{figure}

Besides evolving accurate solutions and wide applicability, another important aspect of Kozax is the computational efficiency. In Fig.~\ref{fig:time}a, we analyze the wall clock time of the three libraries on a symbolic regression task. The symbolic regression task was repeated with increasing number of data points that have to be evaluated. As Kozax can be deployed on both CPU (AMD Genoa 9654, 64 cores) and GPU (NVIDIA A100), the runtime is measured for each device, while PySR and DEAP are only run on CPU. The wall clock time was measured given a fixed number of generations and population size. PySR and DEAP are faster than Kozax running on either CPU or GPU for the smallest dataset. PySR remains faster than Kozax on CPU when the size of the dataset is increased. However, when the number of data points is 100000, PySR and especially DEAP become more inefficient, while Kozax demonstrates high scalability on large datasets or populations when deployed on GPU.

Fig.~\ref{fig:time}b presents the wall clock time of DEAP and Kozax for the objective function optimization task. The objective function optimization was performed only with DEAP and Kozax with the small and large dataset described in Section~\ref{sec: loss}. Here, Kozax was only tested on GPU. Given the small dataset, Kozax runs faster than DEAP with a factor of 30. Even when Kozax is applied to the large dataset, the wall clock time is less than DEAP on the small dataset. Kozax demonstrates highly efficient computation for large datasets and difficult problems.

Figure~\ref{fig:time}a shows that PySR and DEAP are faster than Kozax on running on either CPU or GPU for smaller datasets of symbolic regression. PySR remains faster than Kozax on CPU for the increased size of the dataset. However, when the number of data points is 100000, the possibility to run Kozax on GPU shows to be beneficial, as its runtime is lower than PySR and DEAP. When running on GPU, the computation time of Kozax scales well when more data points or candidates have to be evaluated. Figure~\ref{fig:time}b presents the wall clock time of DEAP and Kozax for the objective function optimization task. Given the small dataset, Kozax runs faster than DEAP with a factor of 30. Even when Kozax is applied to the large dataset, the wall clock time is less than DEAP on the small dataset. Kozax demonstrates highly efficient computation for large datasets and difficult problems.

\section{Discussion}
In this paper, we introduced Kozax, a library for genetic programming (GP) built on JAX. By representing trees as matrices, while following the standard GP algorithm, Kozax parallelizes the evaluation of trees with different structures. This way, the fitness evaluation is sped up compared to evaluating candidate solutions sequentially. Additionally, Kozax runs op GPU, therefore the fitness evaluation scales efficiently to large populations, datasets and difficult problems.

GP has shown promising performance in several domains, such as symbolic regression~\cite{bongard2007automated, cranmer2023interpretable} and symbolic policy optimization~\cite{hein2018interpretable, vries2024discovering}. Most GP libraries are designed specifically for a certain class of problems, however a general framework that can handle arbitrary problem classes was still missing. We developed Kozax to give users much freedom in their problem definition. In the results, we showed that Kozax performs competitively compared to PySR on symbolic regression tasks and outperforms DEAP. Yet, Kozax can also be applied to evolve control problems and objective functions, which is impossible with PySR without using external methods. DEAP can be extended to such applications, but the results showed that DEAP could not evolve accurate solutions efficiently. Besides the scalability, another advantage of Kozax is that it flexibly learns trees for desired functionality.

As demonstrated, the matrix representation of trees in Kozax improves the computation speed of fitness evaluation and reproduction. The representation of the solution space has resemblances to related algorithms like grammar evolution~\cite{o2001grammatical}, Cartesian GP~\cite{miller2015cartesian} and linear GP~\cite{brameier2007basic}. In Kozax, the matrices implicitly adhere to the hierarchical tree structure in the standard GP algorithm. A big advantages of tree-based GP is that only active nodes can be changed during evolution, while grammar evolution, Cartesian GP and linear GP suffer from inefficient evolution of inactive nodes. To our knowledge, Kozax is the first implementation that uses a matrix representation for the standard GP algorithm.

Although Kozax demonstrated efficient evolution of accurate solutions, there are still features that could further improve the applicability of Kozax to complex problems. PySR showed that integrating simplification of expressions throughout evolution and adding constraints of symbolic functions improve the interpretability of the discovered expressions~\cite{cranmer2023interpretable}. Another improvement would be allow for higher-dimensional inputs and outputs in the trees, which relates to strongly typed GP~\cite{montana1995strongly}. Processing higher-dimensional inputs would enable Kozax to be applied to visual data or vector operations. Being able to evolve a tree with multiple outputs allows to learn the same functionality for multiple variables, which could for example evolve compact neural network structures. Extending GP with automatically defined functions (ADF) supports learning useful building blocks that may be included repeatedly in other trees. Especially when evolving multiple trees simultaneously, it would be efficient to reuse functionality in different trees. The additional value of ADFs increases even more when operators with more than two inputs could be evaluated in Kozax, as complex functions structure can be evolved.

In fields such as scientific discovery~\cite{wang2023scientific} and explainable artificial intelligence~\cite{puiutta2020explainable}, it is beneficial to have interpretable white-box models. As GP automatically generates interpretable computer programs, it is a fundamental approach to optimize white-box models. By utilizing GP, the resulting models may provide knowledge about the underlying system, as seen in symbolic regression, or transparency in decision with symbolic control policies. With the development of Kozax, we hope to contribute to the creation of trustworthy artificial intelligence. 

\section{Acknowledgements}
This publication is part of the project ROBUST: Trustworthy AI-based Systems for Sustainable Growth with project number KICH3.L TP.20.006, which is (partly) financed by the Dutch Research Council (NWO), ASMPT, and the Dutch Ministry of Economic Affairs and Climate Policy (EZK) under the program LTP KIC 2020-2023. All content represents the opinion of the authors, which is not necessarily shared or endorsed by their respective employers and/or sponsors.

\printbibliography

@article{koza1994genetic,
  title={Genetic programming as a means for programming computers by natural selection},
  author={Koza, John R},
  journal={Statistics and Computing},
  volume={4},
  pages={87--112},
  year={1994},
  publisher={Springer}
}

@article{schmidt2009distilling,
  title={Distilling free-form natural laws from experimental data},
  author={Schmidt, Michael and Lipson, Hod},
  journal={Science},
  volume={324},
  number={5923},
  pages={81--85},
  year={2009},
  publisher={American Association for the Advancement of Science}
}

@article{bongard2007automated,
  title={Automated reverse engineering of nonlinear dynamical systems},
  author={Bongard, Josh and Lipson, Hod},
  journal={Proceedings of the National Academy of Sciences},
  volume={104},
  number={24},
  pages={9943--9948},
  year={2007},
  publisher={National Acad Sciences}
}

@article{cao2000evolutionary,
  title={Evolutionary modeling of systems of ordinary differential equations with genetic programming},
  author={Cao, Hongqing and Kang, Lishan and Chen, Yuping and Yu, Jingxian},
  journal={Genetic Programming and Evolvable Machines},
  volume={1},
  pages={309--337},
  year={2000},
  publisher={Springer}
}

@article{hein2018interpretable,
  title={Interpretable policies for reinforcement learning by genetic programming},
  author={Hein, Daniel and Udluft, Steffen and Runkler, Thomas A},
  journal={Engineering Applications of Artificial Intelligence},
  volume={76},
  pages={158--169},
  year={2018},
  publisher={Elsevier}
}

@article{vries2024discovering,
  title={Discovering Dynamic Symbolic Policies with Genetic Programming},
  author={de Vries, Sigur and Keemink, Sander and van Gerven, Marcel},
  journal={arXiv preprint arXiv:2406.02765},
  year={2024}
}

@article{jordan2021evolving,
  title={Evolving interpretable plasticity for spiking networks},
  author={Jordan, Jakob and Schmidt, Maximilian and Senn, Walter and Petrovici, Mihai A},
  journal={Elife},
  volume={10},
  pages={e66273},
  year={2021},
  publisher={eLife Sciences Publications, Ltd}
}

@inproceedings{harding2007fast,
  title={Fast genetic programming on GPUs},
  author={Harding, Simon and Banzhaf, Wolfgang},
  booktitle={Genetic Programming: 10th European Conference, EuroGP 2007, Valencia, Spain, April 11-13, 2007. Proceedings 10},
  pages={90--101},
  year={2007},
  organization={Springer}
}

@inproceedings{miller2015cartesian,
  title={Cartesian genetic programming},
  author={Miller, Julian and Turner, Andrew},
  booktitle={Proceedings of the Companion Publication of the 2015 Annual Conference on Genetic and Evolutionary Computation},
  pages={179--198},
  year={2015}
}

@book{brameier2007basic,
  title={Basic Concepts of Linear Genetic Programming},
  author={Brameier, Markus F and Banzhaf, Wolfgang},
  year={2007},
  publisher={Springer}
}

@article{andre1998parallel,
  title={A parallel implementation of genetic programming that achieves super-linear performance},
  author={Andre, David and Koza, John R},
  journal={Information Sciences},
  volume={106},
  number={3-4},
  pages={201--218},
  year={1998},
  publisher={Elsevier}
}

@inproceedings{koza1992automatic,
  title={Automatic programming of robots using genetic programming},
  author={Koza, John R and Rice, James P},
  booktitle={AAAI},
  volume={92},
  pages={194--207},
  year={1992}
}

@book{koza1990genetic,
  title={Genetic programming: A Paradigm for Genetically Breeding Populations of Computer Programs to Solve Problems},
  author={Koza, John R},
  volume={34},
  year={1990},
  publisher={Stanford University, Department of Computer Science Stanford, CA}
}

@software{jax2018github,
  author = {James Bradbury and Roy Frostig and Peter Hawkins and Matthew James Johnson and Chris Leary and Dougal Maclaurin and George Necula and Adam Paszke and Jake Vander{P}las and Skye Wanderman-{M}ilne and Qiao Zhang},
  title = {{JAX}: composable transformations of {P}ython+{N}um{P}y programs},
  url = {http://github.com/jax-ml/jax},
  version = {0.3.13},
  year = {2018},
}

@inproceedings{lange2023evosax,
  title={evosax: Jax-based evolution strategies},
  author={Lange, Robert Tjarko},
  booktitle={Proceedings of the Companion Conference on Genetic and Evolutionary Computation},
  pages={659--662},
  year={2023}
}

@article{fortin2012deap,
  title={DEAP: Evolutionary algorithms made easy},
  author={Fortin, F{\'e}lix-Antoine and De Rainville, Fran{\c{c}}ois-Michel and Gardner, Marc-Andr{\'e} Gardner and Parizeau, Marc and Gagn{\'e}, Christian},
  journal={The Journal of Machine Learning Research},
  volume={13},
  number={1},
  pages={2171--2175},
  year={2012},
  publisher={JMLR. org}
}

@article{cranmer2023interpretable,
  title={Interpretable machine learning for science with PySR and SymbolicRegression.jl},
  author={Cranmer, Miles},
  journal={arXiv preprint arXiv:2305.01582},
  year={2023}
}

@article{wang2025evogp,
  title={EvoGP: A GPU-accelerated Framework for Tree-Based Genetic Programming},
  author={Wang, Lishuang and Wu, Zhihong and Sun, Kebin and Li, Zhuozhao and Cheng, Ran},
  journal={arXiv preprint arXiv:2501.17168},
  year={2025}
}

@inproceedings{bengio1994use,
  title={Use of genetic programming for the search of a new learning rule for neural networks},
  author={Bengio, Samy and Bengio, Yoshua and Cloutier, Jocelyn},
  booktitle={Proceedings of the First IEEE Conference on Evolutionary Computation. IEEE World Congress on Computational Intelligence},
  pages={324--327},
  year={1994},
  organization={IEEE}
}

@inproceedings{burlacu2020operon,
  title={Operon C++ an efficient genetic programming framework for symbolic regression},
  author={Burlacu, Bogdan and Kronberger, Gabriel and Kommenda, Michael},
  booktitle={Proceedings of the 2020 Genetic and Evolutionary Computation Conference Companion},
  pages={1562--1570},
  year={2020}
}

@article{la2021contemporary,
  title={Contemporary symbolic regression methods and their relative performance},
  author={La Cava, William and Burlacu, Bogdan and Virgolin, Marco and Kommenda, Michael and Orzechowski, Patryk and de Fran{\c{c}}a, Fabr{\'\i}cio Olivetti and Jin, Ying and Moore, Jason H},
  journal={Advances in neural information processing systems},
  volume={2021},
  number={DB1},
  pages={1},
  year={2021},
  publisher={NIH Public Access}
}

@inproceedings{topchy2001faster,
  title={Faster genetic programming based on local gradient search of numeric leaf values},
  author={Topchy, Alexander and Punch, William F and others},
  booktitle={Proceedings of the Genetic and Evolutionary Computation Conference},
  volume={155162},
  year={2001},
  organization={Morgan Kaufmann San Francisco, CA}
}

@article{wang2023scientific,
  title={Scientific discovery in the age of artificial intelligence},
  author={Wang, Hanchen and Fu, Tianfan and Du, Yuanqi and Gao, Wenhao and Huang, Kexin and Liu, Ziming and Chandak, Payal and Liu, Shengchao and Van Katwyk, Peter and Deac, Andreea and others},
  journal={Nature},
  volume={620},
  number={7972},
  pages={47--60},
  year={2023},
  publisher={Nature Publishing Group UK London}
}

@article{oussaidene1997parallel,
  title={Parallel genetic programming and its application to trading model induction},
  author={Oussaidene, Mouloud and Chopard, Bastien and Pictet, Olivier V and Tomassini, Marco},
  journal={Parallel Computing},
  volume={23},
  number={8},
  pages={1183--1198},
  year={1997},
  publisher={Elsevier}
}

@inproceedings{nadizar2024naturally,
  title={Naturally interpretable control policies via graph-based genetic programming},
  author={Nadizar, Giorgia and Medvet, Eric and Wilson, Dennis G},
  booktitle={European Conference on Genetic Programming},
  pages={73--89},
  year={2024},
  organization={Springer}
}

@inproceedings{puiutta2020explainable,
  title={Explainable reinforcement learning: A survey},
  author={Puiutta, Erika and Veith, Eric MSP},
  booktitle={International Cross-Domain Conference for Machine Learning and Knowledge Extraction},
  pages={77--95},
  year={2020},
  organization={Springer}
}

@book{kepler1997harmony,
  title={The Harmony of the World},
  author={Kepler, Johannes},
  volume={209},
  year={1997},
  publisher={American Philosophical Society}
}

@article{newton1987philosophiae,
  title={Philosophi{\ae} naturalis principia mathematica (Mathematical principles of natural philosophy)},
  author={Newton, Isaac},
  journal={London (1687)},
  volume={1687},
  number={1687},
  pages={1687},
  year={1987}
}

@book{bonnet1781contemplation,
  title={Contemplation de la Nature},
  author={Bonnet, Charles},
  volume={4},
  year={1781},
  publisher={De l'imprimerie de Samuel Fauche, libraire du roi}
}

@article{goel1971volterra,
  title={On the Volterra and other nonlinear models of interacting populations},
  author={Goel, Narendra S and Maitra, Samaresh C and Montroll, Elliott W},
  journal={Reviews of Modern Physics},
  volume={43},
  number={2},
  pages={231},
  year={1971},
  publisher={APS}
}

@article{sutton1995generalization,
  title={Generalization in reinforcement learning: Successful examples using sparse coarse coding},
  author={Sutton, Richard S},
  journal={Advances in Neural Information Processing Systems},
  volume={8},
  year={1995}
}

@software{gymnax2022github,
  author = {Robert Tjarko Lange},
  title = {{gymnax}: A {JAX}-based Reinforcement Learning Environment Library},
  url = {http://github.com/RobertTLange/gymnax},
  version = {0.0.4},
  year = {2022},
}

@inproceedings{raymond2023fast,
  title={Fast and efficient local-search for genetic programming based loss function learning},
  author={Raymond, Christian and Chen, Qi and Xue, Bing and Zhang, Mengjie},
  booktitle={Proceedings of the Genetic and Evolutionary Computation Conference},
  pages={1184--1193},
  year={2023}
}

@inproceedings{wagner2005heuristiclab,
  title={Heuristiclab: A generic and extensible optimization environment},
  author={Wagner, Stefan and Affenzeller, Michael},
  booktitle={Adaptive and Natural Computing Algorithms: Proceedings of the International Conference in Coimbra, Portugal, 2005},
  pages={538--541},
  year={2005},
  organization={Springer}
}

@article{stanley2002evolving,
  title={Evolving neural networks through augmenting topologies},
  author={Stanley, Kenneth O and Miikkulainen, Risto},
  journal={Evolutionary Computation},
  volume={10},
  number={2},
  pages={99--127},
  year={2002},
  publisher={MIT Press}
}

@inproceedings{tang2022evojax,
  title={Evojax: Hardware-accelerated neuroevolution},
  author={Tang, Yujin and Tian, Yingtao and Ha, David},
  booktitle={Proceedings of the Genetic and Evolutionary Computation Conference Companion},
  pages={308--311},
  year={2022}
}

@article{montana1995strongly,
  title={Strongly typed genetic programming},
  author={Montana, David J},
  journal={Evolutionary Computation},
  volume={3},
  number={2},
  pages={199--230},
  year={1995},
  publisher={MIT Press One Rogers Street, Cambridge, MA 02142-1209, USA journals-info~…}
}

@article{huang2024evox,
  title={EvoX: A Distributed GPU-accelerated Framework for Scalable Evolutionary Computation},
  author={Huang, Beichen and Cheng, Ran and Li, Zhuozhao and Jin, Yaochu and Tan, Kay Chen},
  journal={IEEE Transactions on Evolutionary Computation},
  year={2024},
  publisher={IEEE}
}

@inproceedings{bechtle2021meta,
  title={Meta learning via learned loss},
  author={Bechtle, Sarah and Molchanov, Artem and Chebotar, Yevgen and Grefenstette, Edward and Righetti, Ludovic and Sukhatme, Gaurav and Meier, Franziska},
  booktitle={25th International Conference on Pattern Recognition (ICPR)},
  pages={4161--4168},
  year={2021},
  organization={IEEE}
}

@article{o2001grammatical,
  title={Grammatical evolution},
  author={O'Neill, Michael and Ryan, Conor},
  journal={IEEE Transactions on Evolutionary Computation},
  volume={5},
  number={4},
  pages={349--358},
  year={2001},
  publisher={IEEE}
}

@inproceedings{cano2014gpu,
  title={GPU-parallel subtree interpreter for genetic programming},
  author={Cano, Alberto and Ventura, Sebasti{\'a}n},
  booktitle={Proceedings of the Genetic and Evolutionary Computation Conference Companion},
  pages={887--894},
  year={2014}
}

@article{robilliard2009genetic,
  title={Genetic programming on graphics processing units},
  author={Robilliard, Denis and Marion-Poty, Virginie and Fonlupt, Cyril},
  journal={Genetic Programming and Evolvable Machines},
  volume={10},
  pages={447--471},
  year={2009},
  publisher={Springer}
}

@inproceedings{langdon2008simd,
  title={A SIMD interpreter for genetic programming on GPU graphics cards},
  author={Langdon, William B and Banzhaf, Wolfgang},
  booktitle={European Conference on Genetic Programming},
  pages={73--85},
  year={2008},
  organization={Springer}
}

@article{trujillo2022gsgp,
  title={Gsgp-cuda—a cuda framework for geometric semantic genetic programming},
  author={Trujillo, Leonardo and Contreras, Jose Manuel Mu{\~n}oz and Hernandez, Daniel E and Castelli, Mauro and Tapia, Juan J},
  journal={SoftwareX},
  volume={18},
  pages={101085},
  year={2022},
  publisher={Elsevier}
}

@article{castelli2019gsgp,
  title={GSGP-C++ 2.0: A geometric semantic genetic programming framework},
  author={Castelli, Mauro and Manzoni, Luca},
  journal={SoftwareX},
  volume={10},
  pages={100313},
  year={2019},
  publisher={Elsevier}
}

@article{towers2024gymnasium,
  title={Gymnasium: A Standard Interface for Reinforcement Learning Environments},
  author={Towers, Mark and Kwiatkowski, Ariel and Terry, Jordan and Balis, John U and De Cola, Gianluca and Deleu, Tristan and Goul{\~a}o, Manuel and Kallinteris, Andreas and Krimmel, Markus and KG, Arjun and others},
  journal={arXiv preprint arXiv:2407.17032},
  year={2024}
}

@inproceedings{langdon2010many,
  title={A many threaded CUDA interpreter for genetic programming},
  author={Langdon, William B},
  booktitle={European Conference on Genetic Programming},
  pages={146--158},
  year={2010},
  organization={Springer}
}

@article{sathia2021accelerating,
  title={Accelerating genetic programming using gpus},
  author={Sathia, Vimarsh and Ganesh, Venkataramana and Nanditale, Shankara Rao Thejaswi},
  journal={arXiv preprint arXiv:2110.11226},
  year={2021}
}

@article{chitty2017faster,
  title={Faster GPU-based genetic programming using a two-dimensional stack},
  author={Chitty, Darren M},
  journal={Soft Computing},
  volume={21},
  pages={3859--3878},
  year={2017},
  publisher={Springer}
}





\end{document}